\documentclass[times,twocolumn,final,authoryear]{elsarticle}

\usepackage{prletters}
\usepackage{framed,multirow}
\usepackage{url}
\usepackage{amssymb} \newcommand{\stitle}[1]{\vspace{1ex}\noindent\textup{\textbf{#1}}}
\usepackage{amssymb}
\usepackage{latexsym}
\usepackage{url}
\usepackage{subfigure}
\usepackage[english]{babel}
\usepackage{arabtex}
\usepackage{amsmath}
\usepackage{algorithm}
\usepackage[noend]{algpseudocode}

\begin{document}
\setarab

\vskip1pc

\begin{frontmatter}

\title{Large Vocabulary Arabic Online Handwriting Recognition System}

\author[1]{Ibrahim \snm{Abdelaziz}\corref{cor1}} 
\author[1]{Sherif \snm{Abdou}}
\author[2]{Hassanin \snm{Al-Barhamtoshy}}

\address[1]{Faculty of Computers \& Information, Cairo University, Giza, Egypt}
\address[2]{Faculty of Computing and Information Technology, King Abdulaziz University,Jeddah, Saudi Arabia} 

\begin{abstract}

Arabic handwriting is a consonantal and cursive writing. The position of the character and its context define its shape.
The analysis of Arabic script is
further complicated due to obligatory dots/strokes that are placed above or
below most letters and usually written delayed in order. 
Due to ambiguities and diversities of the different writing styles, recognition systems are generally based 
on a set of possible words called lexicon (vocabulary). When the lexicon is small, 
recognition accuracy is more important as the recognition time is minimal. 
On the other hand, recognition speed as well as the accuracy are both critical issues when handling large lexicons.
Arabic language is rich in morphology and syntax which makes its lexicon large.
Therefore, a practical online handwriting recognition system should be able to handle the large lexicon of the Arabic language with
reasonable performance in terms of both accuracy and time. \\
In this paper, we introduce a fully-fledged Hidden Markov Model (HMM) based system for Arabic online handwriting recognition
that provides solutions for most of the difficulties inherent in recognizing the Arabic script. A new
preprocessing technique for handling the delayed strokes
is introduced. 
We use advanced modeling techniques for building our recognition system from the training data to
provide more detailed representation for the differences between the writing units, 
minimize the variances between writers in the training data, enhance the models discrimination power and have a better representation for
the features space.
The system results are enhanced using an additional
post-processing step to rescore multiple hypothesis of the system result with
higher order language model and cross-word HMM models. The system performance
is evaluated using two different databases covering small and large
lexicons. 
Our proposed system outperforms the state-of-art systems for the small lexicon database. 
Furthermore, it shows promising results (accuracy and time) when supporting large vocabulary with 
the possibility for adapting the models for specific writers to get even better results.
}
\end{abstract}

\begin{keyword}
\KWD Online Handwriting Recognition\sep Arabic\sep Large Vocabulary\sep Adaptive Training\sep Hidden Markov Models\sep Advanced Modeling
\end{keyword}
\end{frontmatter}

\setarab

\section{Introduction}
\label{intro}
The wide spread use of pen-based hand held devices such as PDAs, smartphones, and tablet-PC,
increases the demand for high performance on-line handwritten recognition systems.
This man machine interface method is an alternative for the traditional keyboard with
the advantages of being more easy, friendly, and natural.
\let\thefootnote\relax\footnote{Preprint submitted to Pattern Recognition Letters.}

Automatic Handwritten  Recognition can be classified into  two  types:
on-line  and off-line  recognition. Off-line recognition does not require
a direct interaction with the user. It just applies feature extraction on
the
scanned images of the handwritten text. In on-line recognition, a time
ordered sequence of coordinates (representing  the  movement of the pen ) is
captured and fed to the system as a sequence of 2D-points in real-time, thus
tracking additional temporal data not present in off-line recognition.
Online handwriting  recognition is becoming more and more important in the
modern world due to the spread of hand-held devices. It becomes more challenging
when dealing with cursive language like Arabic.

Arabic text, both handwritten and printed, is cursive. The letters are joined
together along a writing line. In contrast to Latin text, Arabic is written
right to left, rather than left to right. It contains dots and other
small marks that can change the meaning of a word. The shapes of the letters
differ depending on whereabouts in the word they are found. The same letter
at the beginning and end of a word can have a completely different
appearance. Along with the dots and other marks representing vowels, this
makes the effective size of the alphabet about 160 characters. 

The morphology of the Arabic language poses special challenges to 
computational natural language processing systems. 
The Arabic language has large lexicons containing 30,000 to 90,000 words (\cite{wshah2010novel}). 
Research in on-line handwritten word recognition has traditionally concentrated on relatively small lexicons. 
Several Researchers (\cite{abdelazeem2011line,eraqi2011line,ahmed2011line,hosny2011using}) proposed online handwriting
recognition systems for the Arabic language. However, their approaches are developed for small vocabulary (1000 words) and they did not 
aim at solving the challenges imposed by supporting large vocabulary lexicons. 
On the other hand, \cite{biadsy2006online} proposed a HMM-based handwriting recognition system with support for a 40,000 words lexicon.
However, their method for handling the delayed strokes requires the initial detection of the delayed strokes which is a challenging task
by itself. Furthermore, it is sensitive to the writing styles and could misplace the projection of the delayed strokes in the word body. Finally, they tested
their method on a very small dataset consisting of around 3,000 words collected from 10 writers.
Considering the rich morphology and syntax of the Arabic language makes it a must for an effective recognition system to
handle large vocabulary lexicon.

In this paper, we introduce a large vocabulary HMM-based system for online Arabic
handwriting recognition. This system supports a vocabulary size of 64k
unique words which represents 92\% coverage for the Arabic language. Our system is
inspired with the similarity between speech recognition and handwriting
recognition, as both of them can be considered a stochastic process with
sequential nature. Several advanced HMM modeling and training techniques that
are adopted in most state of the art speech recognition systems
are used for building our system models. A novel preprocessing method for handling delayed strokes is presented.
Unlike previous efforts, it solves the delayed strokes problem in most of the writing styles.

\stitle{Our Contributions: } In summary, our specific contributions are:
\begin{enumerate}
 \item A fully-fledged Arabic Online Handwriting system with a support for a large vocabulary (64,000 unique words).
 \item  A novel preprocessing method for handling delayed strokes is presented. Unlike previous efforts,
        it solves the delayed strokes problem in most of the writing styles.
 \item With only few data samples, our models can be adapted for certain writers and have significantly better performance.
 \item Our system outperforms state-of-the-art approaches for small vocabulary. Furthermore, it shows very promising results in terms of 
	both accuracy and time when supporting a large vocabulary.
\end{enumerate}

\begin{figure*}[t]
\centering
  \includegraphics[width=0.9\textwidth,height=4cm]{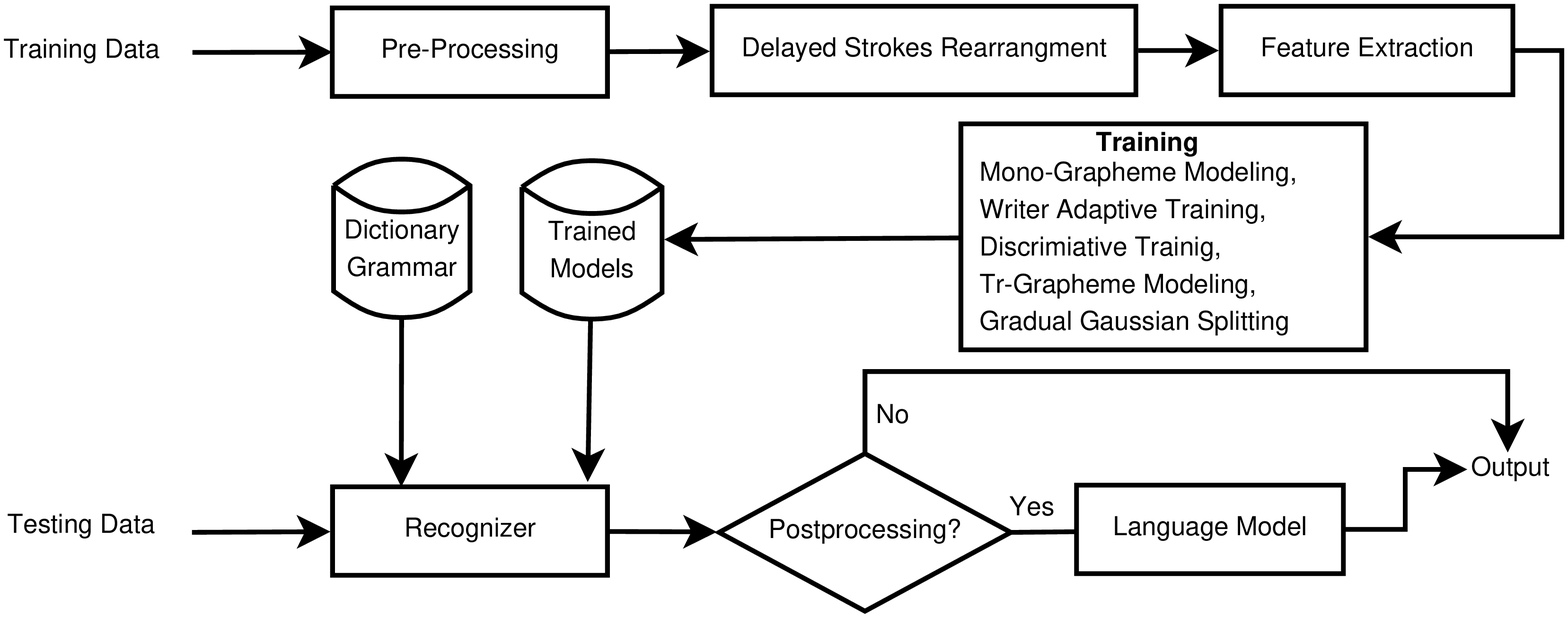}
  \caption{Block Diagram}
  \label{fig:blockDiagram}
\end{figure*}
\stitle{System Overview: }
Figure \ref{fig:blockDiagram} shows the system block diagram. Preprocessing
operations are used to reduce the effect of the handwriting device noise and
the handwriting irregularity.  Then the delayed strokes are rearranged to
match the structure of the HMM model. A new approach for delayed strokes
handling is developed. Our approach overcomes the limitations of the method introduced by \cite{biadsy2006online}.
Our method does a finer projection to avoid the misplacing of the projection points into wrong places in the written word body. An
advantage of our approach is that, it does not require the initial detection of the delayed strokes as all the strokes of the input are handled similarly.
Several features are extracted from the handwriting signal and are used to train the HMM models. In the recognition phase, our trained models are used with the
application dictionary by a decoding engine to select the best words that match the user input. Optionally, the output of the first recognition phase can be
passed to another post-processing step to rescore multiple hypothesis of the system results with a higher order language model

\stitle{Paper Organization: } The paper is organized as follows:
In Section \ref{sec:related}, we describe the
related work. Section \ref{featurePrep} describes the
pre-processing, feature extraction and the delayed strokes rearrangement
steps. Section \ref{HmmSection} describes the HMM models structure and
training procedure in addition to the post-processing phase. The system
evaluation using small and large databases is introduced in section
\ref{sysEval}. Section \ref{conc} includes the final conclusions and a plan
for the future work.

\section{Related Work}
\label{sec:related}
Early research on online Arabic handwriting recognition focused on the
recognition of isolated characters. \cite{el1989line} proposed a
method for the recognition of handwritten Arabic characters drawn on a
graphic tablet using writer independent features and Freeman-like chain code.
\cite{kharma2001novel} proposed the use of a mapping for the handwritten
characters to normalize the orientation, position, and size of the input
pattern.  \cite{mezghani2002line} investigated a method based on Kohonen
maps and their  corresponding  confusion matrices which serve to prune the
error-causing nodes, and to combine them consequently. \cite{al2005efficient}
proposed an efficient structural approach for recognizing on-line Arabic
handwritten digits based on the changing signs of the slope values to
identify and extract the primitives.

To recognize larger units, \cite{almuallim1987method} proposed a
structural recognition method for cursive Arabic handwritten words by
segmenting them into strokes. These strokes are classified using their
geometrical and topological properties then they are combined into a string
of characters that represents the recognized word. \cite{alimi1997evolutionary} developed an
online writer dependent system to recognize Arabic cursive words based on
neuro-fuzzy approach. \cite{elanwar2007simultaneous}  proposed a system to recognize
online Arabic cursive handwriting based on rule-based method to perform
segmentation and recognition of word portions in an unconstrained cursive
handwritten document using dynamic programming. \cite{daifallah2009recognition}
developed an on-line Arabic handwritten recognition system based on an
arbitrary stroke segmentation algorithm followed by segmentation enhancement,
consecutive joint connections and segmentation point locating.

The structural-based approaches are based upon the idea that character shape
can be described in an abstract fashion without paying too much attention to
the shape variations that necessarily occur during the execution of that
plan. These approaches try to segment the input pattern before applying
recognition for the produced segments. Consequently any error in the
segmentation phase is unrecoverable and would affect the accuracy of the
final recognition result.  A better alternative was proposed by using HMM
models which are doubly stochastic models that proved to achieve good
performance for sequences recognition. Using HMM models the pattern
segmentation and recognition can be achieved simultaneously using an
integrated search technique such as Viterbi or A-star.

\cite{khorsheed2003recognising} have successfully used HMM models for the
recognition of off-line handwritten Arabic script. In their approach, word
level HMM is composed of smaller interconnected models that represent the
character level models. Each character model is a right-to-left HMM.
Structural features are extracted from overlapping vertical windows that
scans the input pattern sequentially from right to left with same direction
as the model structure. 

Recently, several efforts (\cite{abdelazeem2011line,eraqi2011line,ahmed2011line,hosny2011using}) proposed HMM-based
online handwriting recognition systems. However, their approaches are developed
for small vocabulary (1000 words) and they did not aim at solving the challenges imposed
when supporting large vocabulary lexicons.

\stitle{Handling Delayed Strokes}
When dealing with on-line handwriting, the right-to-left order of writing is
not guaranteed. Usually writers tend to use some delayed strokes by moving
backward to add some diacritics. In Arabic, there are 17 characters out of 28 are written
with delayed strokes. So the percentage of characters with delayed strokes is 60\% and will be higher if we included
the different diacritics.
The delayed strokes make disturbance in the order of the writing sequences. It
results in a mismatch with the expected order of the input sequence for the HMM
model according to the constraint of the right-to-left model structure. To
deal with this challenge, four different solutions were proposed in the literature.

\begin{figure}
\centering
  \includegraphics[width=0.37\textwidth]{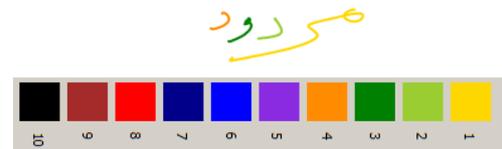}
\caption{Sample Word}
\label{fig:sample}      
\end{figure}

Figure \ref{fig:sample} includes a color legend that is used to show the writing order of different strokes in a word.
The colors describe the order on which the different strokes are written. For example, this sample has four strokes,
the first one is written in yellow while the second one is written in white green and so on.

In the first approach by \cite{abdelazeem2011line}, the delayed strokes are totally discarded from the handwriting in the preprocessing
phase. This method could not be employed effectively since the information
that makes letters different from others is the number and position where the
dots are located. Eliminating delayed strokes will cause a tremendous
ambiguity, particularly when the letter body is not written clearly.
Furthermore, some Arabic letters have a similar shape of composition with
some letters, such as: the letter(s) <s> has a similar shape to the three
letter shapes <bty--> (b + t + y) (Without dots).

\begin{figure}
  \centering
  \subfigure{
       \includegraphics[width=0.37\columnwidth]{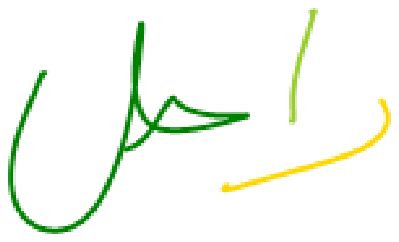}
   }
   \subfigure{
       \includegraphics[width=0.37\columnwidth]{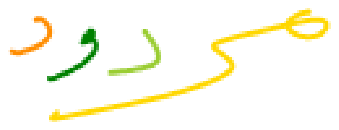}
   }
 \caption{Case 1: Overlapped or small letters can be mis-detected as delayed strokes.}
     \label{fig:overlapped}
 \end{figure}

The second approach is introduced by \cite{ha1993unconstrained}.  Delayed strokes are detected in the
preprocessing phase and then used in a post-processing phase to differentiate
between ambiguous words. The detection of the delayed strokes is by itself a
challenging task and the errors in this preprocessing step can result in
discarding segments form the main body of the handwritten words. For example,
in Figure \ref{fig:overlapped}, there are no delayed strokes. However, letters are totally overlapped
with each others, hence, they can be detected as delayed strokes.

\begin{figure}
  \centering
  \subfigure[Delayed strokes written right to left after all letters]{
       \label{fig:conn1}
       \includegraphics[width=0.37\columnwidth]{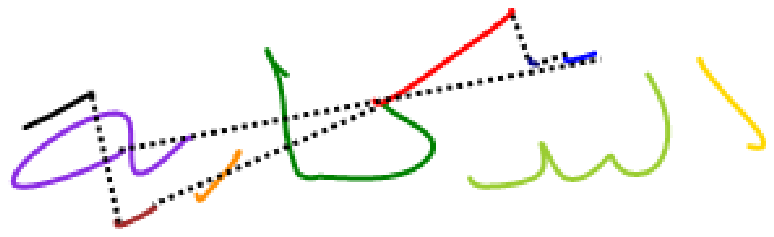}
   }
   \subfigure[Delayed strokes written right to left intermingled with bodies]{
       \label{fig:conn1}
       \includegraphics[width=0.37\columnwidth]{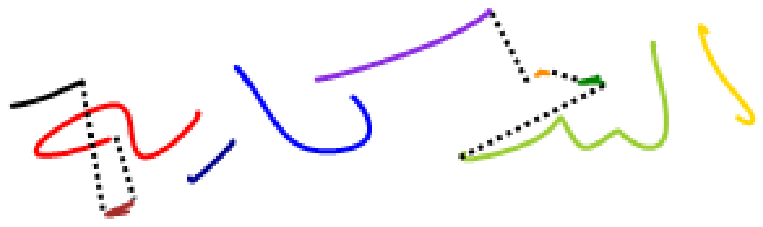}
   }
 \caption{Case 2: Connecting delayed strokes to the end of the word can result in different sequences.}
     \label{fig:conn}
 \end{figure}

\begin{figure}
  \centering
  \subfigure[Written after all letters]{
       \label{fig:7a2e2a1}
       \includegraphics[width=0.37\columnwidth]{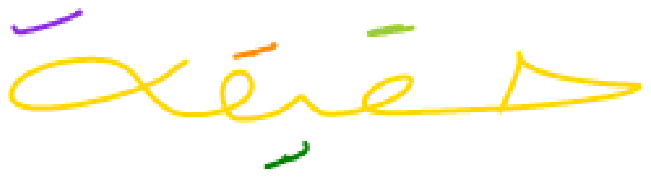}
   }
   \subfigure[Intermingled with letters]{
       \label{fig:7a2e2a2}
       \includegraphics[width=0.37\columnwidth]{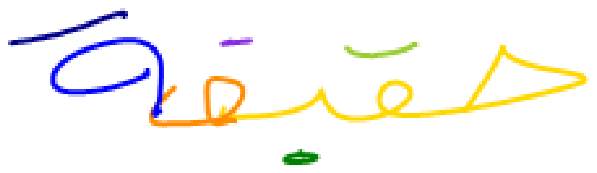}
   }
 \caption{Case 3: Different styles to write delayed strokes, therefore enumerating all possible permutations is not practical.}
     \label{fig:7a2e2a}
 \end{figure}

\setarab
The third approach is introduced by \cite{starner1994line,hu2000writer}; it keeps the delayed strokes with special manipulation.
In \cite{starner1994line}'s approach, the end of a word is connected to the delayed strokes with a
special connecting stroke. The special stroke indicates that the pen was
raised and results in a continuous stroke sequence for the entire handwritten
sentence. Clearly, as shown in Figure \ref{fig:conn}, the order used to write the
delayed strokes change the shape of the whole word greatly. We can see that the same word
with different orders of writing delayed strokes have two different shapes.
Others approaches, like \cite{hu2000writer}, treat delayed strokes as special
characters in the alphabet. So, a word with delayed strokes is given
alternative spellings to accommodate different sequences where delayed
strokes are drawn in different orders. But these two approaches are not
practical as Arabic words may contain many delayed strokes. These methods
will dramatically increase the hypothesis space, since words should be
represented in all of their handwriting permutations. For example: the word <
Al.hqyqT> "the truth" contains 8 dots, thus, 8! representations would be
required. As an example, Figure \ref{fig:7a2e2a} show two different styles for
writing the same word. In Figure \ref{fig:7a2e2a1}, the writer wrote all the letters bodies
then he wrote all delayed strokes. However, Figure \ref{fig:7a2e2a2} shows a different styles for
writing the same word where writing delayed strokes is intermingled with writing the letters bodies.
This shows that this approach is infeasible as it has to cover all possible handwriting representations of each word.

\begin{figure}
\centering
  \includegraphics[width=0.20\textwidth]{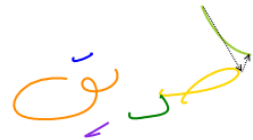}
\caption{Case 4: Delayed Strokes has minimum overlap with the letter body, therefore it can be projected into a wrong place}
\label{fig:fadiIssue}      
\end{figure}

A fourth practical solution to handle delayed strokes is proposed by \cite{biadsy2006online}. The authors project
the delayed strokes inside its related letter body by vertically projecting the first point of the delayed stroke into the
overlapped letter body. The last point of the delayed stroke is connected
to the following point in that letter body. This approach does not require
any restrictions on the order of writing the delayed strokes which makes it
practical but still has two shortcomings. Firstly, its requirement for the
initial detection of the delayed strokes with possibilities of
miss-detections. Secondly, there are cases where the delayed strokes appear
before or after the word-part body where the delayed strokes will be connected
to the closest word-part body. For example, Figure \ref{fig:fadiIssue} shows a sample for this scenario.
The shown word is /Tryq/ "road" which start with the letter /TAH/. The delayed stroke overlap with the
letter body and some part of it is written before the body itself.
Hence, we can see that this approach will project the delayed stroke
before the body itself. As a result, the system might confuse this letter with the two letters <lS-> as they
have the same shape. Moreover, as shown in Figure \ref{fig:overlapped},
the projection technique will harm regular characters that are not delayed strokes because they are totally overlapped.

\section{Data Preparation and Acquisition}
\label{featurePrep}
\subsection{Preprocessing}

 \label{preproc} The goals of the preprocessing
phase are: reduce/remove imperfections caused by acquisition devices, smooth
the irregularity generated by inexperienced writers having an erratic
handwriting and minimize handwriting variations irrelevant for pattern
classification which may exist in the acquired data. The preprocessing
operations used in our system are:

\begin{itemize}
\item \textbf{Removing Duplicated Points} : Duplicated points are
removed by checking whether the coordinates of any two points are the
same, If so only one of them is kept.

	\item	\textbf{Interpolation}: Applying linear interpolation to add
any missing points caused by variation of writing speed (\cite{huang2007preprocessing}).
	\item \textbf{Smoothing}: To eliminate hardware imperfections and
trembles in writing each point is substituted with the weighted average
of its neighboring points (\cite{kavallieratou2002unconstrained}).
	\item \textbf{Re-sampling}: Due to the variation in writing speed,
the acquired points are not distributed evenly along the stroke
trajectory. This operation is used to get a sequence of points which is
equidistant (\cite{kavallieratou2002unconstrained}).
	\item	\textbf{De-hooking}: To remove the hooks that may appear with
sensitive pens at the beginning or end of the strokes due to inaccuracies
in rapid pen-down/up detection and erratic hand-motion
\end{itemize}

\subsection{Delayed Strokes Rearrangement}
\label{subsec:delayed}
The main harm of the delayed strokes is that they result in the scattering of
the character components which does not match the expected sequence of the
HMM model. So the motivation for our solution was to reorder the online
strokes so the closer ones, in the geometric domain, come as successors in
the time domain. But we found that the reordering operation is not only
enough since some strokes can have several characters and the ideal order may
need to insert some delayed strokes in the middle of those long strokes. So
we decided to segment the strokes into small segments then do the reordering
operation on those small segments. At the end of each segment, a geometric
condition is checked to make sure if a delayed stroke needs to be inserted.
After doing all insertions needed, small segments are grouped again together
if it is originally from the same stroke and no insertions happened between
them. This way, we had the flexibility to do a finer reordering that allowed
moving the delayed strokes as much as possible to their ideal location. When
we applied that algorithm to our data, we managed to solve the delayed
strokes problem in more than 96\% of the cases. Even for the redundant
multiple copies of the characters, their harmful effect was minimized. The
delayed strokes reordering algorithm is presented Algorithm
\ref{delStrokeAlg}. The input to the algorithm is the set of the captured strokes
from the handwriting text and a number that defines the size of a stroke segment (line 1); 
the rearranged strokes are added to $OutputInk$. Initially, all strokes are marked as not seen before then 
the algorithm loops through the whole strokes set (line 4-21) and try to reorder the delayed ones.
If the stroke is not used (line 5-6), the stroke is segmented into small segments of the given size (line 7).
After that, the algorithm loops through the whole set of segments while considering all the other input strokes (line 9-20).
It checks if the current segment needs to be inserted in another stroke (line 17-19); if so, the segment order is considered and the algorithm
reorder both stroke and the segment. Finally, after considering all strokes and segments, it returns the final ordered strokes.

\begin{algorithm}\scriptsize
\caption{Delayed Strokes Rearrangement}
\label{delStrokeAlg}

\begin{algorithmic}[1]
\Procedure{RearrangeStrokes}{$Strokes$: Array of input Strokes, $N$: NumberOfStrokes, $S$: Segment Size}
    \State {$OutputInk$: Ordered Strokes}
    \State {Mark all strokes in $Strokes$ as not used}
    \For {$StrokesCounter$ =1 to $N$}
      \If {$Strokes[strokesCounter]$ is used}
	  \State {continue}
      \EndIf
      \Statex
      \State {$Segments=$\\}
      ~~~~~~\Call {SegmentStroke}{$Strokes[strokesCounter]$, $S$}
      \Statex
      \For {$SegmentsCounter = 1$ to $Segments.Size$}
	  
	  \For{$StrokesCounter2 = StrokesCounter+1$ to $N$}
		\If {$Strokes[strokesCounter2]$ is used}
		    \State {continue}
		\EndIf
		\State {$fPtStr=$\\}
		~~~~~~~~~~~~\Call {GetFarthestPoint}{$Strokes[StrokesCounter2]$}
		
		\Statex
		
		\State {$fPtSeg=$\\}
		~~~~~~~~~~~~\Call {GetFarthestPoint}{$Segments[SegmentsCounter]$}
		\Statex
		\If {$FPtStr.x$ $>$ $FPtSeg.x$}
			\State {Add $Strokes[StrokesCounter2]$ to $OutputInk$}
			\State{Mark $Strokes[StrokesCounter2]$ as used}
		\EndIf
	  \EndFor
      \State {Add $Segments[SegmentsCounter]$ to $OutputInk$}
      \EndFor
      \State{Mark $Strokes[StrokesCounter]$ as used}
    \EndFor

    \State Return $OutputInk$
\EndProcedure
\end{algorithmic}
\end{algorithm}

 \begin{figure}
  \centering
  \subfigure[Original Ink]{
       \label{fig:conn1}
       \includegraphics[width=0.4\columnwidth]{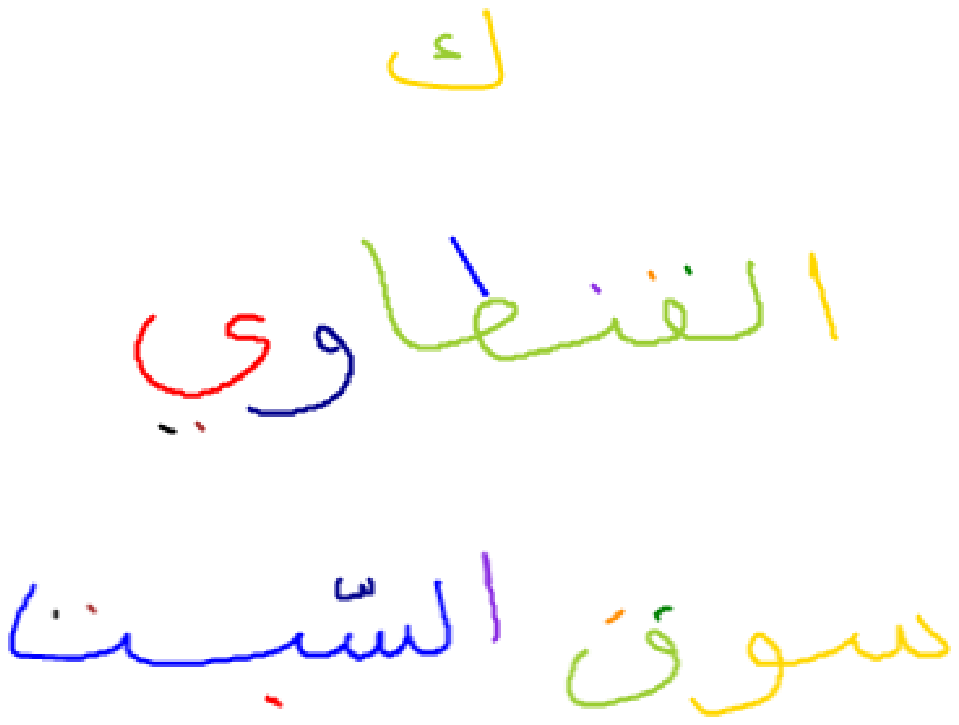}
   }
   \subfigure[Rearranged Ink]{
       \label{fig:conn1}
       \includegraphics[width=0.4\columnwidth]{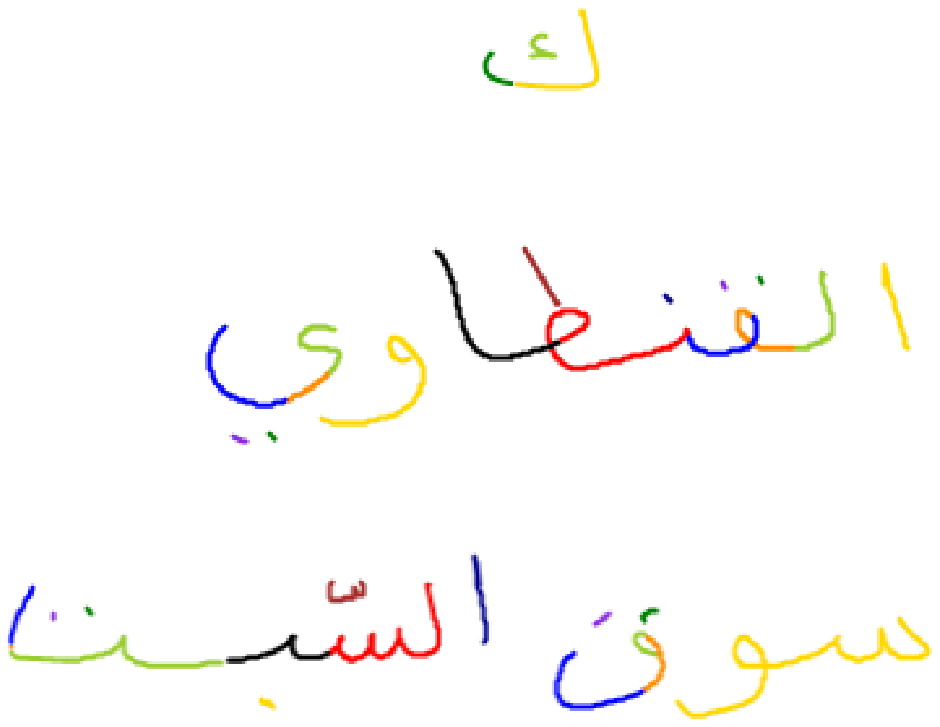}
   }
 \caption{Examples of delayed strokes rearrangement using our method.}
     \label{fig:delayedEx}
 \end{figure}

Figure \ref{fig:delayedEx} shows three examples of delayed stroke
rearrangement. The legend located on the right side shows the order in which
the strokes are written. The first sample shows how delayed strokes are handled in
case of a single letter KAF <k> which has a delayed stroke HAMZA. The delayed
stroke will be inserted in its correct order in the middle of character body.
In the second sample, for the original ink, we can see that the second written stroke
colored with white green contains 4 delayed strokes. After rearrangement, this
stroke is divided into sub strokes in order to have the delayed strokes
inserted at their proper location.

\begin{figure}
  \centering
  \subfigure{
       \includegraphics[scale=0.3,width=0.3\columnwidth]{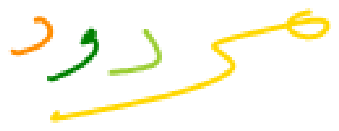}
   }
   \subfigure{
       \includegraphics[scale=0.3,width=0.3\columnwidth]{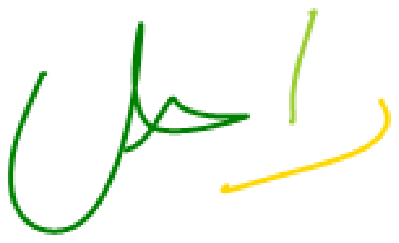}
   }
   \subfigure{
       \includegraphics[scale=0.3,width=0.3\columnwidth]{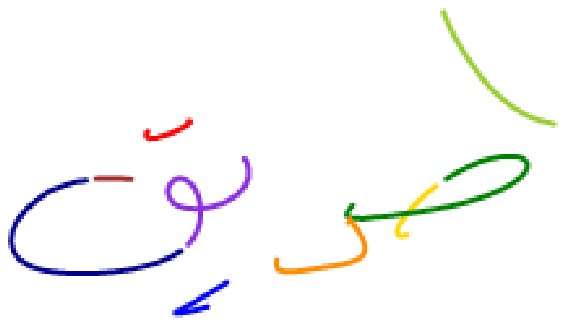}
   }
   \subfigure{
       \includegraphics[scale=0.3,width=0.3\columnwidth]{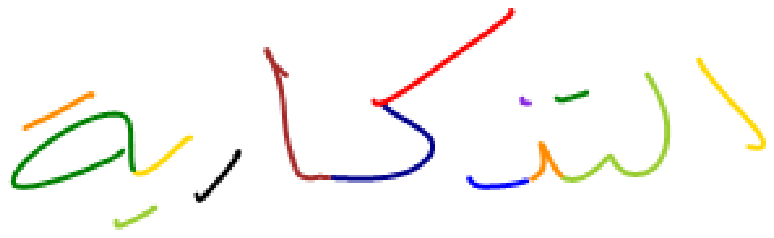}
   }
   \subfigure{
       \includegraphics[scale=0.3,width=0.3\columnwidth]{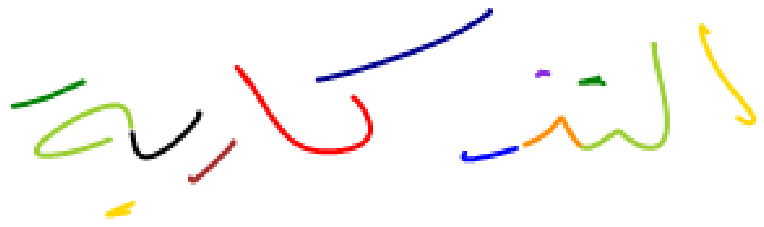}
   }
    \subfigure{
       \includegraphics[scale=0.3,width=0.3\columnwidth]{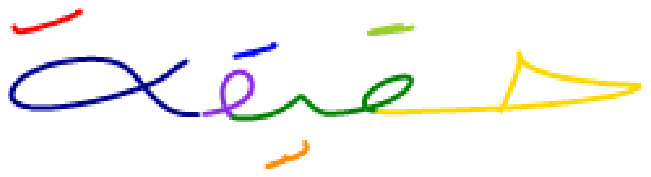}
   }
   \subfigure{
       \includegraphics[scale=0.3,width=0.3\columnwidth]{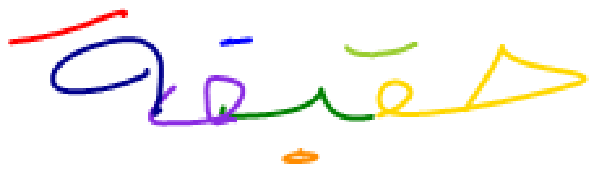}
   }
 \caption{Our rearrangement method addresses all cases.}
     \label{fig:delayedEx2}
 \end{figure}

To further show the effectiveness of our approach, we apply it on all the different
cases discussed above. As we can see, it managed to reorder all the delayed strokes and insert
them in their correct location. Figure \ref{fig:delayedEx2} shows our results.
Notice that, since we are grouping the strokes back after segmentation,
if the algorithm did not detect any delayed strokes, the output from our algorithm will be exactly the same
as the input (see the first two examples).

\subsection{Feature Extraction}
\label{featureEx} In our system we investigated many features and found the
best set of features are as follows:
\subsubsection{Chain Code}
Chain coding is one of the most widely used methods for boundary
description, \cite{wulandhari2008evolution}. This code follows the boundary in counter clockwise
manner and keeps track of the direction as we go from one contour pixel to
the next. A 32-directional chain code is used in our system.
\subsubsection{Curliness}
Curliness C(t)  is a feature that describes the deviation from a straight
line in the vicinity  of (x(t),  y(t)). It is based on the ratio of the
length of the trajectory and the maximum side of the bounding box (\cite{jaeger2001online}):

\begin{center}
$C(t)$ = $\frac{L(t)}{max(\delta{x},\delta{y})}-2$
\label{eq:curliness}
\end{center}

where L(t) denotes the length of the trajectory in the vicinity  of
(x(t),  y(t)), i.e., the sum of lengths of all line segments. $\delta{x}$ and
$\delta{y}$ are the width and height of the bounding box containing all
points in the vicinity of (x(t),  y(t)).  According to this definition, the
values of curliness are in the range [-1;N-3]. However, values greater than 1
are rare in practice.

\subsubsection{Aspect Ratio}
The aspect of the trajectory is a feature which characterizes the
height-to-width ratio of the bounding box containing the preceding and
succeeding points of (x(t), y(t)).  It is described  as a single value A(t):

\begin{center}
$A(t)$ = $\frac{2\delta{y}}{\delta{x} + \delta{y}}-1$
\label{eq:aspect}
\end{center}

Where $\delta{x}$ and $\delta{y}$ are the width and height of the bounding
box containing all points in the vicinity of (x(t),  y(t)).

\subsubsection{Writing  Direction}
The local writing direction at a point (x(t),  y(t))  is described using the
cosine and sine functions as follows:

\begin{center}
$cos\alpha{t}=\frac{\delta{x}(t)}{\delta{s}(t)}$ \\
$sin\alpha{t}=\frac{\delta{y}(t)}{\delta{s}(t)}$
\label{eq:writingDir1}
\end{center}

where $\delta{s}(t), \delta{x}(t)$ and $\delta{y}(t)$ are defined as follows:

\begin{center}
$\delta{s}(t)$=$ \sqrt{\delta{x^{2}}(t) + \delta{y^{2}}(t)}$ \\
$\delta{x}(t)$=$x(t-1)-x(t)$ \\
$\delta{y}(t)$=$y(t-1)-y(t)$
\end{center}

\subsubsection{Curvature}
The curvature of a curve at a point is a measure of how sensitive its tangent
line is to moving that point to other nearby points. The curvature at a point
(x(t),  y(t)) is represented by the cosine and sine of the angle defined by
the following sequence of points : (x(t  - 2), y(t - 2)), (x(t),  y(t)),
(x(t  + 2), y(t + 2)). Strictly  speaking, this signal does not represent
curvature but the angular difference signal. Curvature would be 1/r , of a
circle touching and partially fitting  the curve, with radius r. Cosine and
sine can be computed using the values of the writing  direction :

\begin{center}
$cos\beta{t} = cos \alpha{t - 1} * cos\alpha{t + 1} + sin\alpha{t - 1} * sin\alpha{t + 1}$, \\
$sin\beta{t} = cos\alpha{t - 1} * sin\alpha{t + 1} - sin\alpha{t - 1}
* cos\alpha{t + 1}$
\end{center}

\subsubsection{Baseline and Zones}
This feature represents a vertical reference position for the characters and
words in a handwriting sample. In our system it is
determined using traditional histogram method by projecting the writing
tracing points of a word or line of text onto a vertical line. The baseline
is detected using the maximal peak in that histogram (\cite{huang2007preprocessing})

After detecting the baseline, then the sample is divided into three zones
upper, middle and lower according to its position from the baseline. 

\subsubsection{Loop detection}
\begin{figure}
\centering
  \includegraphics[width=0.25\textwidth]{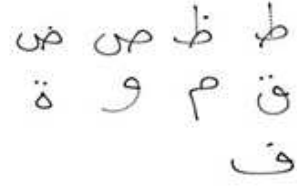}
\caption{Arabic letters with loops}
\label{fig:loops}      
\end{figure}

This is a Boolean feature, which indicate whether the current point is part
of a loop or not. Figure \ref{fig:loops} show Arabic characters containing
loops.

\subsubsection{Hat feature}
This feature indicates whether the current point is part of a delayed stroke
or not. (i.e. the strokes that has been reordered using the previously
described strokes reordering algorithm).

\subsubsection{Extended Features}
After geometric normalization, some extended sequences are derived from the
basic function set. In our system, four dynamic sequences have been used as
extended functions (\cite{jul08biometrics}), namely:

\begin{itemize}
	\item Path-tangent angle
				\begin{center}
				$\theta_{n}$ = $\arctan {y_{n} / x_{n}}$
				\label{eq:paththeta}
				\end{center}
	\item Path velocity magnitude
				\begin{center}
				$v_{n}$ = $\sqrt{x_{n}^{2} + y_{n}^{2}}$
				\label{eq:velmag}
				\end{center}
	\item Log curvature radius:
				\begin{center}
				$\rho_{n}$ = $\log{1/k_{n}} = \log{v_{n}/ \theta_{n}}$
				\label{eq:curvrad}
				\end{center}
			where $k_{n}$ is the curvature of the position trajectory and
log is applied in order to reduce the range of function values.
	\item Total acceleration magnitude:
				\begin{center}
				$a_{n}$ = $\sqrt{t_{n}^{2} + c_{n}^{2}}$
				\label{eq:logcurv}
				\end{center}
			where $t_{n}$ = $v_{n}$ and $c_{n}$ =$v_{n} . \theta_{n}$
			are respectively the tangential and centripetal acceleration
components of the pen motion.
\end{itemize}

\section{HMM Modeling of Handwriting}
\label{HmmSection}
\begin{figure}
\centering
  \includegraphics[width=0.35\textwidth]{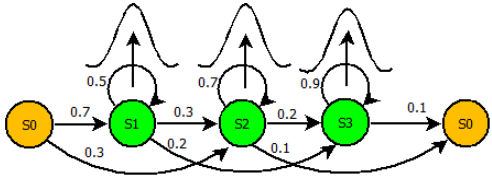}
\caption{HMM model sample }
\label{fig:hmmSample}      
\end{figure}

The proposed system is based on Hidden Markov Models
(HMM).  The HMM is a finite set of states, each of which is associated with a
(generally multidimensional) probability distribution. Transitions among the
states are governed by a set of probabilities called transition
probabilities. Figure \ref{fig:hmmSample} shows a sample HMM model.
Arabic  contains  28 different  letters,  but  as these letters
are position  dependent  it will map  to  103 different  shapes.  In our
proposed system, we have 115 different models. These models include  Arabic
letters  with  their  different  shapes  (103 models), 10 English digits
(0-9),  Arabic  MAD  symbol
and English Capital V letter. This
last two symbols were required for one of the evaluation databases we use (ADAB database).
Also we built models for all the punctuation symbols.

In our system, we use left to right HMM model with different number of states
per model according to how complex the model shape is.
We use variable number of states varying from three to nine states.
Three-States Models are the simplest models that consists of only a single
straight stroke like the digit One \emph{1} and the Arabic letter <A> .
Models with five states are more complex than the previous one as they contain
either two strokes or they shape contain multiple transitions from horizontal
to vertical and so on. Examples are <b-->, <n-->, <y--> and <d>.
When shapes are getting more complex in
terms of the number of strokes and the shape complexity, we model the
characters with more states. For example, <t>, <f>, <lA> and <-H> are
modeled with seven states, whereas <s>, <k>, <S> and <q> are modeled
with nine states.

\begin{figure}
\centering
  \includegraphics[width=0.5\textwidth]{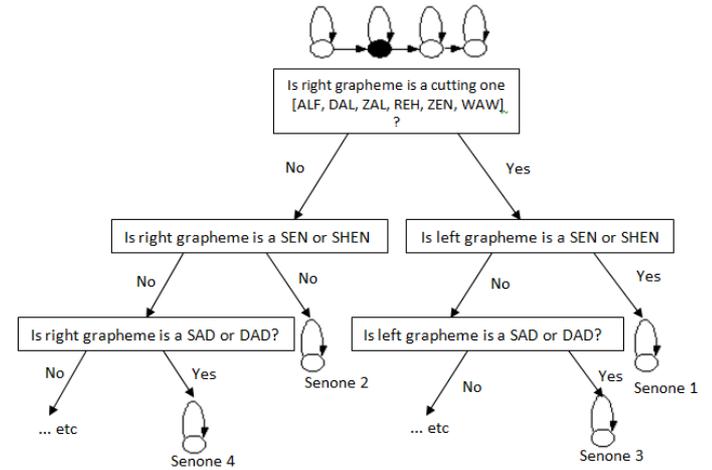}
\caption{clustering questions}
\label{fig:cluster}      
\end{figure}

Initially we built a mono-grapheme system which is based on the 115 different
models mentioned above (position-dependent) using the Maximum Likelihood (ML)
training (\cite{jul08biometrics}) to maximize the probability of the training samples
generated by the model. Then we expanded this initial model to a more
sophisticated HMM model which is tri-graphemes context-dependent model. The
tri-grapheme is a context dependent grapheme unit that considers both the
preceding and following graphe\-mes; for example the letter <m> in word part
<bmd> is different from word part <lmd> though both of them is in the
intermediate position. The tri-grapheme model expansion enables the precise
modeling of the letters shape but with the price of the large increase of the
models numbers. In our Arabic handwriting system with 28 different mono
letters this would require $(28)^{3}$ models. With this large number of
models usually we don't have enough database to train them. In our Arabic
handwriting system we found that the required database to train these 20k
models would be in the size of 8 million words while our training database
included only 150k words. In order to deal with the problem of data
insufficiency, we decided to cluster the HMM states to reduce the number of
the trained models. We used a clustering technique based on decision trees.
It is based on asking questions about the left and right contexts of each
tri-grapheme and clusters together the states that have similar context. The
questions that we used for the models clustering were derived from an
analysis of the Arabic letters shapes and the different handwriting styles.
For example one of the questions that we used ask about the cutting letters
(<a> /ALF/, <d> /DAL/, <r> /REH/, <z> /ZEN/ and <w> /WAW/ and /ZAL/) which are the letters that have to be
followed with the starting position letters. Also we clustered all the
similar characters in shape such as \emph{SEEN} and \emph{SHEEN}, \emph{SAD} and \emph{DAD} ..etc.
Figure \ref{fig:cluster} shows part of the decision tree that we used in our
system.
\setarab

\subsection{Writer Adaptive  Training}
\begin{center}
\begin{figure}
  \includegraphics[width=0.5\textwidth]{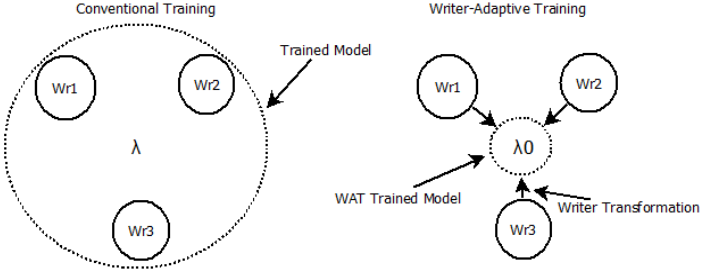}
\caption{Writer Adaptive Training}
\label{fig:wat}      
\end{figure}
\end{center}

To train a robust writer independent handwriting system the training database
should be collected from large number of writers. An inherent difficulty of
this approach is that the resulting statistical models have to contend with a
wide range of variation in the training data caused by the inter-writer
variability. The features distributions will exhibit high variance and hence
high overlap among the different grapheme units which may result in diffused
models with reduced discriminatory capabilities. In speech recognition
systems, Speaker A-daptive training (SAT) was developed to compensate for
speaker differences during acoustic model training (\cite{anastasakos1996compact,anastasakos1996compact}). Each
speaker's training data is linearly transformed so that it more closely
resembles the training data for a prototype speaker. In this way, the models
are made more precise, because the Gaussian doesn't have to model
inter-speaker variability-instead; inter-speaker variability is handled by a
separate speaker normalization step, see figure \ref{fig:wat}.

Similar to the SAT training technique, a Writer Ada-ptive Training (WAT)
technique is employed in our handwriting recognition system. We used
Constrained Maximum Likelihood Linear regression (CMLLR) to adapt each
training writer to the writer-independent model. CMLLR is a feature
adaptation technique that estimates a set of linear transformations for the
features. The effect of these transformations is to shift the feature vector
in the initial system so that each state in the HMM system is more likely to
generate the adaptation data (\cite{young1997htk}). Then the adapted training data for
each writer was used to train a new writer-independent model. Figure \ref{fig:wat}
illustrates this idea. This type of training reduced the variation by moving
all writers towards their common average. Results of the testing data sets
that we used to evaluate our system have shown significant increase in the
system recognition accuracy after applying the WAT approach.
\subsection{Discriminative Training}
Historically, the predominant training technique for HMM has been the Maximum
Likelihood Estimation (MLE).  The MLE technique gives optimal estimates only
if the model correctly represents the stochastic process, an infinite amount
of training data is available and the true global maximum of the likelihood
can be found. In practice, none of the above conditions is satisfied. This
was the motivation for using discriminative training. The discriminative
learning schemes such as Maximum Mutual Information (MMI), Minimum Word Error
(MWE), Minimum Phone Error (MPE) and Minimum Classification Error (MCE) has
recently gained tremendous popularity in machine learning since it makes no
explicit attempt to model the underlying distribution of data and instead it
directly optimizes a mapping function from the input data samples to the
desired output labels. Therefore, in discriminative learning methods, only
the decision boundary is adjusted without forming a data generator in the
entire feature space (\cite{zhou2009discriminative}).

In our system we increased the discrimination power of our models using a
discriminative training scheme similar to the minimum phone error (MPE)
approach. The training procedure is the same but we replaced the training
unit to be a grapheme rather than the phoneme unit.  The training criteria
are:
\begin{center}
\begin{equation}
F_{MGE}(M) = \sum_{H} \frac{P^{k}(O | H,M)P(H)}{\sum_{\breve{H}}P^{k}(O | \breve{H},M)P(\breve{H})}A(H, H_{ref})
\label{eq:de}
\end{equation}
\end{center}
Where O is the observation sequence of the training utterance, M is a model
parameters and H and $H_{ref}$  both denote possible hypotheses of the
training data. $A(H;H_{ref} )$ is the grapheme accuracy of the hypothesis H
given the reference $H_{ref}$. It equals the number of reference graphemes
minus the number of errors. Two sets of lattices are needed: a lattice for
the correct transcription of each training file, and a lattice derived from
the recognition of each training file. These are called the numerator
and denominator  lattices respectively. Then the optimality criterion
of equation \ref{eq:de} is used. In our system we name it the Minimum
Grapheme Error (MGE), a one that tries to reduce the number of grapheme
errors in the final result. Evaluation results show significant improvement
of our system models after applying the discriminative training approach.

\subsection{Gaussian Mixtures}
In the final training step the Gaussian PDFs are converted into Mixture
Gaussian PDFs. This process is done by splitting the Gaussians to increase
their coverage for the features space. That process has to be done slowly,
because any mixture Gaussian with number of mixtures larger than 1 suffers
from spurious and undesirable global optimum parameter settings; i.e., if you
try to learn a 128-component mixture Gaussian all at once without proper
initialization, the training algorithm will learn a set of parameters that
work really well for the training data and really badly for anything else,
usually including at least one nearly-zero variance parameter. In order to
avoid these effects, in our system training procedure we split the Gaussians
gradually, e.g., going from one Gaussian to two, then to four, and so on,
checking the variances at each step to make sure no variance parameter is
getting too small (\cite{gales2001adaptive}). We applied this gradual Gaussians splitting
approach in our system and achieved much better performance than training all
the Gaussians at once as shown in our system evaluation results.

\subsection{Post-processing}
In our HWR system we use a multi-pass decoding approach. Ideally, a decoder
should consider all possible hypotheses based on a unified probabilistic
framework that integrates all knowledge sources such as the HMM handwriting
models and the language models. It is desirable to use the most detailed
models, such as context-dependent models and high order n-grams in the search
as early as possible. The Arabic language is extremely rich in inflections.
As a result, a large dictionary is required to provide practical coverage for
the language. When the explored search space becomes unmanageable, due to the
increasing size of vocabulary or highly sophisticated knowledge sources,
search might be infeasible to implement. A possible alternative is to perform
a multi-pass search and apply several knowledge sources at different stages
in the proper order to constraint the search progressively. In the initial
pass, the most discriminant and computationally affordable knowledge sources
are used to reduce the number of hypotheses. In subsequent passes,
progressively reduced sets of hypotheses are examined, and more powerful and
expensive knowledge sources are then used.

In our system we use two passes. In the first pass we use the most
discriminant and computationally affordable knowledge sources which are
word-internal tri-grapheme HMM model with bi-gram language model. The output
of this first pass is a word lattice which represents a search space with
reduced sets of hypotheses. This lattice includes several alternative words
that were recognized at any given time during the search. It also typically
contains other information such as the time segmentations for these words,
and their HMM and language scores. In the second pass, we rescore this lattice
with more powerful and expensive knowledge sources which are cross word
tri-grapheme HMM model and a fifth-gram language model. To build this language model,
we used a text corpus collected from crawling Aljazeera news website (\cite{aljazeera}). We collected around 700 MB
of text containing 132 million words, each word is four characters on average. The language model is built
using SRI language modeling toolkit (\cite{srilm}) with its default parameters. The lattice error
rate is typically much lower than the word error rate of the single best
hypotheses produced for each sentence. The multi-pass systems implementation
is a successful approach to break the tie between speed and accuracy. With
this approach it is possible to improve decoding accuracy with minor
degradation in decoding speed.

\section{Experimental Results}
\label{sysEval}

\subsection{Small Vocabulary database}
In the first evaluation the HWR system is evaluated against
other state of art HRW systems. Only one international event was found for
Arabic handwriting evaluation. This is the ICEDAR conference that is based on
the ADAB database. This database was developed in cooperation between the
Institut fuer Nachrichtentechnik (IfN) and the Research group on Intelligent
Machines (REGIM).  The database consists around 20K samples written by
more than 170 different writers, most of them selected from the narrower
range of the National school of Engineering of Sfax (ENIS). The ADAB-database
is divided to 4 sets. Details about the number of files, words, characters,
and writers for each set 1 to 4 are shown in Table \ref{adabStat}.

\begin{table}\scriptsize
\centering
\caption{ADAB Database Characteristics}
\label{adabStat}      
\begin{tabular}{|l|l|l|l|l|}
\hline
Set	& Files	& Words	& Characters	& Writers \\\hline
1 & 5037 & 7670 & 40500 & 56 \\ \hline
2 & 5090 & 7851 & 41515 & 37 \\  \hline
3 & 5031 & 7730 & 40544 & 39 \\   \hline
4 & 4417 & 6671 & 35253 & 41 \\  \hline
\end{tabular}
\end{table}

\cite{el2011line} held the first competition on ADAB database
at 10th International Conference on Document Analysis and Recognition
(ICDAR), three data sets were provided for training (sets 1,2 and 3) and set
4 was used for testing the systems. Later in 2011, new test sets (set f and s) were used in
ICDAR 2011 competition. However, these test sets are not publicly available and as a result we could not use them for
evaluating our system.
The results of set 4 for all the competing systems are shown in Table \ref{icdarset4}.
\begin{center}
\begin{table}\scriptsize
\centering
\caption{ADAB Set4 results (ICDAR 2009)}
\label{icdarset4}      
\begin{tabular}{|l|l|l|l|l|}
\hline
System	& Method & Top 1	& Top 5	& Top 10\\     \hline
MDLSTM-1	& NeuralNet	& 95.70 & 98.93 & 100\\ \hline
MDLSTM-2	& NeuralNet	& 95.70 & 98.93 & 100\\ \hline
VisionObjects-1	& NeuralNet&	98.99 & 100 & 100\\  \hline
VisionObjects-2	& NeuralNet	& 98.99 & 100 & 100\\          \hline
REGIM-HTK	& HMM	& 52.67 & 63.44 & 64.52\\            \hline
REGIM-Cv	& VC	& 13.99 & 31.18 & 37.63\\             \hline
REGIM-CvHTK	& HMM\&VC & 38.71 & 59.07 & 69.89\\           \hline
\end{tabular}
\end{table}
\end{center}

Our system evaluation using ADAB is shown in Table \ref{ourset4}. We
experimented 5 different groups of preprocessing operations which are:

\begin{itemize}
	\item No Preprocessing: Raw data.
	\item Preprocessing 1: Delayed Strokes Reordering.
	\item Preprocessing 2: Delayed Strokes Reordering, Resampling and
Interpolation.
	\item Preprocessing 3: Delayed Strokes Reordering, Resampling,
Interpolation and Smoothing.
	\item Preprocessing 4: Delayed Strokes Reordering, Resampling ,
Interpolation, Smoothing, Duplicate Points Removal and Dehooking.
\end{itemize}

\begin{table}\scriptsize
\centering
\caption{System evaluation for the ADAB Database}
\label{ourset4}      
\begin{tabular}{|l|l|l|l|}
\hline
System&Top 1&Top 5&Top 10\\
\hline
Mono-Grapheme + No Preproc&2.15&8.08&14.49\\        \hline
Mono-Grapheme  + Preproc 1&92.66&97.85&98.50\\      \hline
Mono-Grapheme  + Preproc 2&93.52&97.92&98.39\\      \hline
Mono-Grapheme  + Preproc 3&93.79&97.92&98.60\\      \hline
Mono-Grapheme + Preproc 4&94.43&98.52&98.92\\       \hline
+Writer Adaptive Training&94.83&98.56&  98.91\\     \hline
+Discriminative Training&95.98&98.42&99.17\\        \hline
+Tri-Grapheme&96.18&98.90&99.13\\                   \hline
+Gradual Gaussians&97.13&99.11&99.40\\              \hline
\end{tabular}
\end{table}

From the results shown in Table \ref{ourset4}, we can see how promising our system
performance compared to the state of the art systems.  Results show that the
Delayed Strokes Reordering is an essential operation in the system. Also the
other utilized preprocessing operations have provided absolute 1.8\%
improvement in the system accuracy. The used advanced training techniques
provided another 2.2\% improvements in accuracy. It is worth mentioning here
that all the experiments in Table \ref{ourset4} are using the same feature set
defined in Section \ref{featureEx}

\begin{table}\scriptsize
\centering
\caption{ALTEC database statistics}
\label{altecStat}      
\begin{tabular}{|l|l|l|}
\hline
 	& Total Number &	Unique entries \\
\hline
Words & 152680 & 39945 \\ \hline
PAWs  & 325477 & 14740 \\  \hline
Pages  & 4512 & - \\       \hline
Writers & 1000 & - \\       \hline
\end{tabular}
\end{table}

\subsection{Large Vocabulary}
Our second concern was evaluating our system in a large vocabulary task. We
evaluated the system using the ALTEC Arabic Handwriting (ALTECOnDB) database
(\cite{altecEsole}). This database contains handwriting samples from 1000 different
writers comprised of men and women from various professional backgrounds,
qualifications, and ages. Each writer was asked to write 4 pages that
contains 200 words on average. The written text was selected from the
Gigaword Arabic text database. A 30K sentences were selected from that
database with 99\% coverage of the paws of the Arabic language. Table
\ref{altecStat} show the statistics of the ALTECOnDB database.

For system testing  we used the ALTEC-AH test set. This test set is collected
by 16 writers. Each writer wrote 11 pages with average 750 words. The Out Of
Vocabulary (OOV) ratio for this test according to a 64k dictionary is 8.3\%.
Detailed statistics are shown in Table \ref{testAHstats}.

\begin{table}\scriptsize
\centering
\caption{ALTEC-AH test set statistics}
\label{testAHstats}      
\begin{tabular}{| l | l |}
\hline
 Number of writers & 16 \\     \hline
Number of pages	& 176\\        \hline
Number of lines	& 1717\\        \hline
Number of Words	& 12853\\       \hline
OOV words	& 1066\\ \hline
\end{tabular}
\end{table}

We generated a writer-dependent models from the writer-independent ones using the CMLLR technique discussed before.
The writer-dependent models are created for the different writers in the test set by splitting the test set ALTEC-AH into 
adaptation part and testing part. For each writer in the test data, we
used only 4 pages for the model adaptation. The rest 7 pages are used for evaluating the writer-dependent models.

In this experiment, we show the performance of our system, both writer-independent and dependent models, using two passes:
\begin{itemize}
  \item Pass 1: uses the same trained models used for evaluating ADAB database but with a a larger dictionary of size 64K words. 
  \item Pass 2: The output of the first pass is a lattice which includes several alternative words that were recognized at 
  any given time during the search. It also typically
contains other information such as the time segmentations for these words,
and their HMM and language scores. In the second pass, we rescore this lattice with a high-level (five-grams) 
linguistic model to improve the performance of the first pass.
\end{itemize}

Table \ref{ourtestAH} shows the evaluation results of our system for the ALTEC-AH test set.
\begin{table}\scriptsize
\centering
\caption{System evaluation - ALTEC-AH}
\label{ourtestAH}      
\begin{tabular}{|l|l|l|}
\hline
 	System	& Pass1 Accuracy	& Pass2 Accuracy \\
\hline
Writer-Independent & 68.76 & 80.07 \\ \hline
Adapted models  & 79.40 & 87.47 \\    \hline
\end{tabular}
\end{table}
We can see that our results are very promising. After the second pass of the writer-independent models,
the system's accuracy increased from 68\% to 80\%. 
For the writer-dependent models, the system acheived an accuracy of 79.4\% in pass 1 which increased to 87.5\%  
after using the high-order language model.
Notice that, after adapting the system models to match the writing style and characteristics of the system users, our system could boost the
accuracy to 87.5\% and this was achieved with an amount of adaptation data
less than 200 words per writer. The streamed output results of the system, i.e the
immediate partial results without waiting for writing the whole sentence, are
only 79\% for the adapted system and 68\% for the writer independent system
which is still not a practical accuracy. If we exclude the OOV words from our
evaluation results the in-vocabulary accuracy is 87\% for the
writer-independent system and 95\% for the adapted system. We did not find any
references for reported results on comparable large vocabulary Arabic
handwriting systems to compare our system against them.

\subsection{Runtime}
In this experiment, we report the average time it takes our system
to recognize a sample. All experiments are run using a Lenovo z560 laptop with 4G RAM and 2.53GHz Intel core i5 processor.
The laptop is running 64-bit Microsoft Windows 7.

Table \ref{tab:set4_times} shows the time it takes our system on average to recognize ADAB Set4 samples. A sample can be
a single or few words. We report the average time it takes our system to output the top 1,5 and 10 results respectively.
As we can see, when supporting small vocabulary, our system is almost real-time. It takes less than a second to produce the full top 10
matches of a given sample.

\begin{table}\scriptsize
\centering
\caption{Small Vocabulary Running Time (ADAB Database)}
\label{tab:set4_times}      
\begin{tabular}{|l|l|l|l|}
\hline
 &\multicolumn{3}{c}{Average time per sample(seconds)}\\ \hline
Database &Top 1&Top 5&Top 10\\                   \hline
ADAB Set 4 &0.448&0.9372&0.923\\                      \hline
\end{tabular}
\end{table}

In Table \ref{tab:altec_times}, we report the running time of our system when working on large vocabulary.
As expected, as the supported vocabulary size increases, our system takes more time (2.2 seconds) to recognize a test sample.
Although, Pass 2 takes very small time (0.15 seconds) to rescore
the produced lattice from Pass 1, it managed to get significantly more accurate results as shown in Table \ref{ourtestAH}.

\begin{table}\scriptsize
\centering
\caption{Large Vocabulary Running Time (ALTECOnDB)}
\label{tab:altec_times}      
\begin{tabular}{|l|l|l|}
\hline
 &\multicolumn{2}{c}{Average time per word(seconds)}\\     \hline
Database &Pass 1&Pass 2\\               \hline
ALTEC-AH &2.2&0.15\\                           \hline
\end{tabular}
\end{table}

\section{Conclusion}
\label{conc} We proposed a system for large-vocabulary Arabic online Handwriting
recognition that provides solutions for most of the difficulties inherent in recognizing the Arabic script.
A new approach for handling the delayed strokes is introduced which avoids the drawbacks of the previously introduced methods
in literature. Our system is based on Hidden Markov Models and trained with advanced modeling techniques adopted by 
speech recognition systems such as context dependent modeling, speaker adaptive
training, discriminative training and Gaussians mixtures splitting.
The system results are enhanced using an additional post-processing step to rescore multiple hypothesis of the system result with
higher order language model and cross-word HMM models.\\

Our HWR system outperforms state-of-the-art research efforts when evaluated using a data set with a small vocabulary.
Furthermore, when tested on a large vocabulary database (ALTECOnDB), the results we obtained are very 
promising in terms of both accuracy and running time. The advantage of our system is its simple structure, and its adopted
models are based on mature technology for sequential data modeling. With only few data samples, the writer independant models can be adapted
for a certain writer to acheive better accuracy.
In the future work, we plan to expand the system vocabulary up to half million words
to reach 99\% coverage of the Arabic language. This would require the
investigation of using some of the fixed search decoding techniques such as
finite state decoders.

\bibliographystyle{model2-names}

\end{document}